\documentclass{article}
\usepackage{spconf,amsmath,graphicx}

\usepackage{enumitem}
\setlist{nosep, leftmargin=14pt}

\usepackage{mwe} 

\usepackage[caption=false,font=small]{subfig}
\usepackage{booktabs}

\usepackage{xcolor}
\usepackage[final]{changes}
\setdeletedmarkup{\textcolor{red}{\sout{#1}}}
\setaddedmarkup{\textcolor{blue}{\uline{#1}}}

\usepackage{lipsum} 

\usepackage{amssymb}

\usepackage{subfig}

\usepackage{multirow}

\usepackage[detect-all=true,separate-uncertainty=true, multi-part-units=single,range-phrase={-},product-units = single]{siunitx}
\DeclareSIUnit{\pixel}{px}
\DeclareSIUnit{\fps}{fps}

\title{6D Robotic OCT Scanning of Curved Tissue Surfaces}
\name{Suresh Guttikonda$^{*, 1}$\thanks{* These authors contributed equally.} \quad Maximilian Neidhardt$^{*, 1,3,4}$ \quad Vidas Raudonis$^{2,4}$ \quad Alexander Schlaefer$^{1,3,4}$}
\address{
    $^{1}$ Institute of Medical Technology and Intelligent Sys., Hamburg University of Technology, Germany \\
    $^{2}$ Faculty of Electrical and Electronics Eng., Kaunas University of Technology, Lithuania \\
    $^{3}$ Interdisciplinary Competence Center for Interface Research, Hamburg, Germany \\
    $^{4}$ SustAInLivWork Center of Excellence, Kaunas, Lithuania \\
}

\usepackage{footmisc}
\begin{document}

%
\maketitle

\begin{abstract}
\replaced{Optical coherence tomography (OCT) is a non‑invasive volumetric imaging modality with high spatial and temporal resolution. Imaging larger tissue regions requires moving the OCT probe to cover the target area. 
Handheld scanning approaches rely on overlapping volumes for stitching, whereas, robotic scanning approaches often restrict motion to pure translations to avoid full hand–eye calibration — a challenging problem given the OCT probe’s small field of view. These registration- or translation-based stitching strategies, however, struggle when imaging curved tissue surfaces. We propose a fiducial marker enabling full six‑degree‑of‑freedom hand–eye calibration for a robot‑mounted OCT probe. Our method produces highly repeatable transform estimates and, in experiments on two phantom surfaces, enables consistent robotic scanning of large, curved geometries. Because the approach does not depend on image‑based registration, it avoids error accumulation along scanning paths and outperforms conventional purely translational robotic scanning.} {Optical coherence tomography (OCT) is a non-invasive volumetric imaging modality with high spatial and temporal resolution. For imaging larger tissue structures, OCT probes need to be moved to scan the respective area. For handheld scanning, stitching of the acquired OCT volumes requires overlap to register the images. For robotic scanning and stitching, a typical approach is to restrict the motion to translations, as this avoids a full hand-eye calibration, which is complicated by the small field of view of most OCT probes. However, stitching by registration or by translational scanning are limited when curved tissue surfaces need to be scanned. We propose a marker for full six-dimensional hand-eye calibration of a robot mounted OCT probe. We show that the calibration results in highly repeatable estimates of the transformation. Moreover, we evaluate robotic scanning of two phantom surfaces to demonstrate that the proposed calibration allows for consistent scanning of large, curved tissue surfaces. As the proposed approach is not relying on image registration, it does not suffer from a potential accumulation of errors along a scan path. We also illustrate the improvement compared to conventional 3D-translational robotic scanning.}
\end{abstract}

\section{INTRODUCTION}

Optical coherence tomography (OCT) is a light based image modality with high spatial and temporal resolution. In the last decade, several clinical OCT imaging systems have been introduced for scanning of small structures, e.g., in ophthalmology for imaging the retina of the eye~\cite{pircher2017review}. Scanning larger areas requires scanning motion, e.g., along vessels to identify plaque in intravascular OCT (IVOCT)~\cite{gessert2018automatic}. Furthermore, OCT probes have been proposed to scan tissue surfaces and estimate elastic tissue properties in larger areas~\cite{neidhardt2024deep}. However, typically volumetric scanning with an OCT scan head is restricted to a rather small field of view (FOV) of approximately \qtyproduct{10 x 10 x 2.7}{\milli\meter} along the lateral and depth axes, respectively. Hence, imaging of larger organs or surfaces is limited and can restrict the applicability of OCT.

\begin{figure}[t]
    \replaced{\centerline{\includegraphics[width=\linewidth, height=60mm]{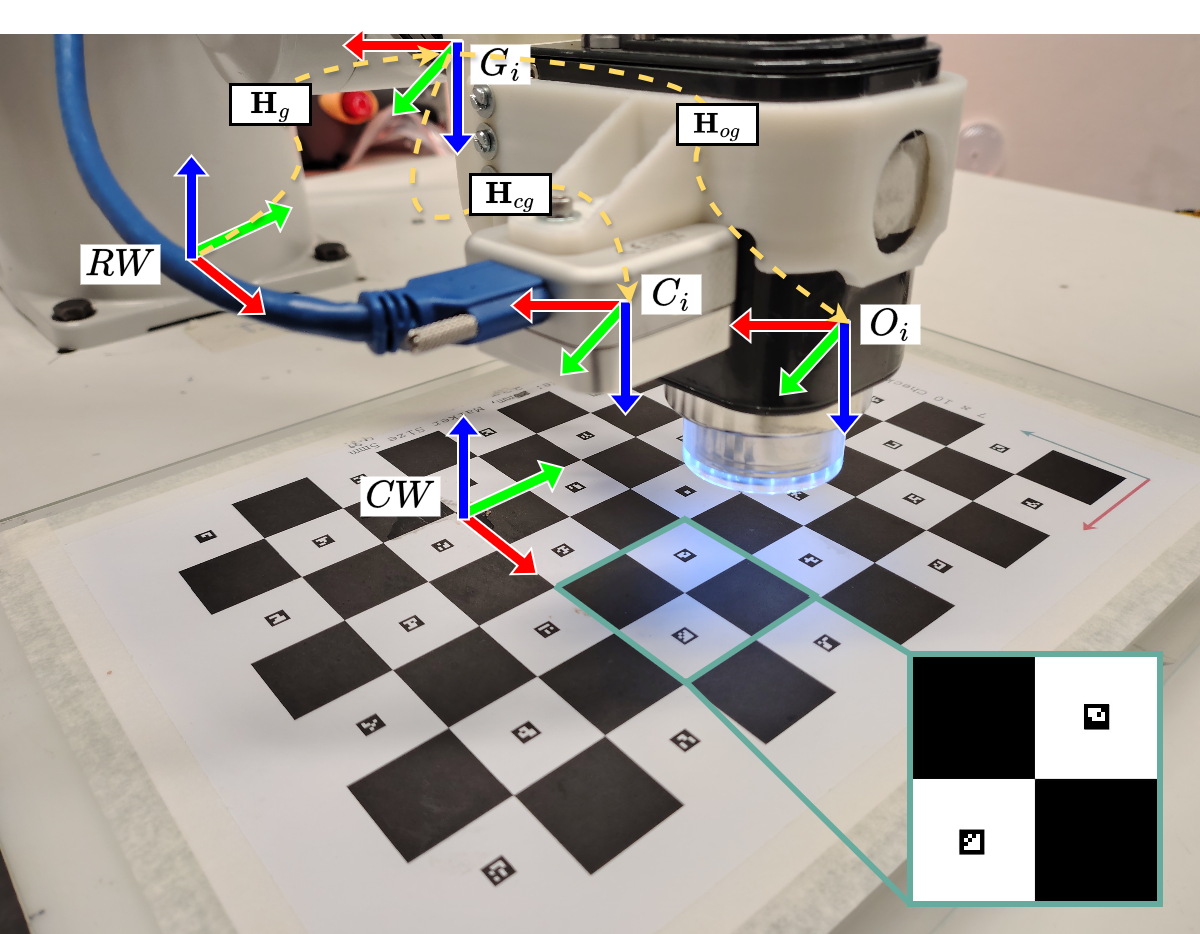}}}{\centerline{\includegraphics[width=\linewidth]{Setup_Suresh.drawio.png}}}
    \caption{\textbf{OCT Surface Scanning:} An OCT scan head ($O$) and a RGB-D camera ($C$) are mounted to a robot ($G$). A custom calibration pattern ($CW$) is used to calibrate the camera and OCT scan head to the robot world coordinate frame ($RW$).}
    \label{img:OCT_realsense_abb_setup}
\end{figure}

Robots are ideal for enlarging the FOV by driving a scan head close to the surface. After scanning, the volumetric data can be stitched together based on the recorded robot positions or features extracted from the volumetric data. Previous works move the robot in a grid pattern along the lateral axis of the OCT volume. During scanning, the surface is detected and the robot position is adjusted accordingly to keep the surface inside the limited FOV~\cite{sprenger2021automated, lotz2024large}. However, the surface topography can change abruptly while driving between grid points, which can limit surface detection as the scan object might not be visible inside the FOV. Hence, only small scan steps are feasible to ensure sufficient overlap between adjacent volumes while the orientation of the robot's end-effector (EEF) is fixed. Thereby, volumes can be stitched together based on the recorded robot positions~\cite{sprenger2021automated}. Lotz et al. similarly perform a raster scan with a robot mounted OCT scanner. They image larger structures by extracting en-face projections from volumetric data and stitch panoramic 2D images based on image features. These approaches do not require a calibration between OCT and robot reference frame. However, scanning of complex surface structures is restricted as the orientation of the EEF needs to remain fixed, which limits the scan motion, particularly for curved surfaces.

To enable complex robot trajectories during scanning a 6D calibration between the OCT scan head and the Robot reference frame is necessary, referred to as hand-in-eye calibration. Commonly, checkerboard phantoms with lateral dimensions above \SI{100}{\milli\meter} are scanned with cameras, e.g., RGB-D cameras. Similarly, large OCT scans can also be acquired by scanning these phantoms in a grid pattern and stitching individual volumes together based on the recorded robot poses \cite{prakash2025portable}. This approach has also been demonstrated to be feasible for 1-dimensional depth scan probes attached to the robot~\cite{rajput2016high}. However, here the marker detection itself requires stitching of OCT volumes, i.e., the robot's pose changes during acquisition of the calibration data.

To this end, we propose a custom calibration pattern containing features to estimate the transformations between RGB-D camera w.r.t to the robot base and OCT scan head w.r.t to the robot base. We do not stitch volumes during calibration data acquisition and thereby mitigate calibration errors arising from volume stitching. The calibration phantom contains conventional ArUco markers and a checkerboard pattern as depicted in Fig.~\ref{img:OCT_realsense_abb_setup}. During calibration data acquisition, we acquire OCT volumes encompassing all ArUco marker features. We systematically evaluate the calibration accuracy and demonstrate full 6D scanning on phantoms and soft tissue.



\begin{figure}
  \centering
  \begin{tikzpicture}
    \node[inner sep=0pt] (bg) at (0,0) {\includegraphics[trim=0cm 0cm 0cm 0cm, clip, width=\linewidth]{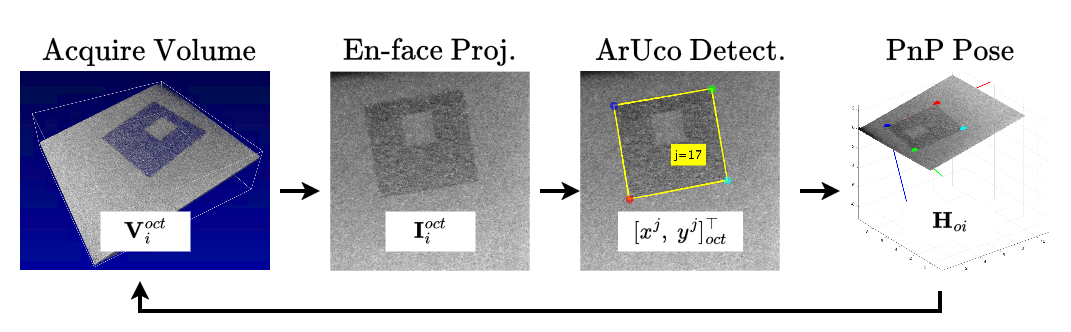}};
  \end{tikzpicture}
  \caption{Pipeline to compile a calibration dataset.}
  \label{img:aruco_marker_detection_workflow}
\end{figure}

\vspace{-3mm}
\section{MATERIAL AND METHODS}

\subsection{Experimental Setup}
Our experimental setup is depicted in Fig.~\ref{img:OCT_realsense_abb_setup}. We mount a custom 3D printed adapter to the EEF of a robot arm (IRB 120, ABB) and attach a RGB-D camera (Intel RealSense D405, Intel) and a OCT scan head (PR3-13-AL, Thorlabs). The scan head is connected to an OCT system (Telesto I, Thorlabs) with a temporal scanline rate of $\SI{91.1}{\kHz}$ and a depth axis resolution of \SI{5.2}{\micro\meter} in air. All acquired OCT volumes have a FOV of $\SI{10}{\mm} \times \SI{10}{\mm} \times \SI{2.66}{\mm}$ and a spatial resolution of $\SI{512}{\pixel} \times \SI{512}{\pixel} \times \SI{512}{\pixel}$ along the lateral and depth axis, respectively. Acquisition time for each OCT volume is $\sim \SI{5.04}{\sec}$, and we compensate for spatial distortions. 
All acquired RGB-D images have a spatial resolution of $\SI{848}{\pixel} \times \SI{480}{\pixel}$. We estimate the camera intrinsics $\mathbf{K}_{c}$ and distortion coefficients ${\kappa}_{c}$ using established camera calibration methods. We design a custom calibration phantom to estimate the orientation of the OCT scan head and the RGB-D camera w.r.t. the robot base. The phantom contains a black and white checkerboard with $10 \times 7$ squares with an edge length of $\SI{10}{mm}$ and ArUco markers of size $\qtyproduct{5 x 5}{mm}$ that are printed at the center of each white square. We refer to this phantom as ChArUco calibration pattern.

\begin{figure}
    \centering
    \subfloat[]{
        \begin{tikzpicture}
        \node[anchor=south west,inner sep=0] (img) at (0, 0){\includegraphics[height=50mm]{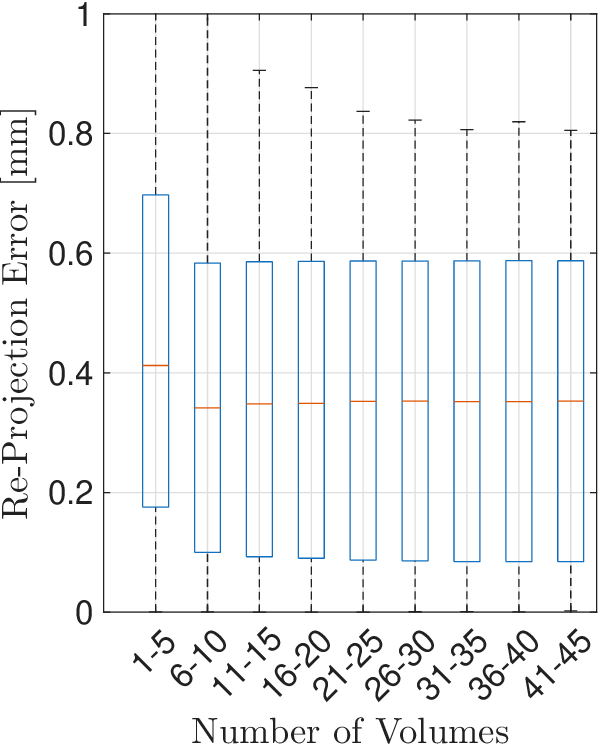}}; 
        \end{tikzpicture} 
        \label{img:translationError_Calibration}
        }
    \subfloat[]{
        \begin{tikzpicture}
        \node[anchor=south west,inner sep=0] (img) at (0, 0){\includegraphics[height=50mm]{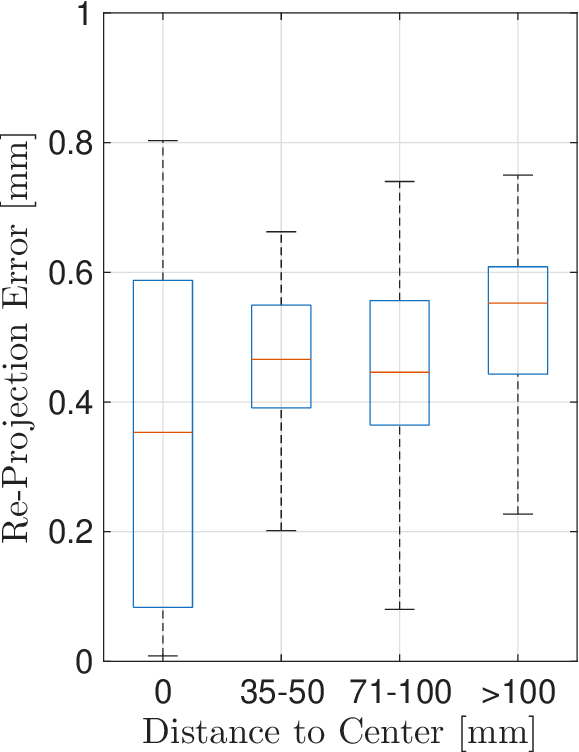}}; 
        \end{tikzpicture} 
        \label{img:distanceError_Calibration}
        }
    \caption{\textbf{Calibration Accuracy:} The re-projection error vs (a) distinct OCT-EEF pose pairs acquired at ArUco marker ($j = 17$)  (b) ArUco markers ($j \neq 17$) at various distances from center ArUco marker $j=17$.}
        \label{img:calibAcc}
\end{figure}

\vspace{-3mm}
\subsection{Notation}

Let $\mathbf{H} = [\mathbf{R} | \mathbf{t}] \in \mathbb{R}^{4 \times 4}$ denotes the homogeneous matrix representation of the 3D orientation matrix $\mathbf{R} \in SO(3)$ and 3D position vector $\mathbf{t} \in \mathbb{R}^{3}$. For any $i, k \in 1 \cdots N$ configurations, let $O_i$, $C_i$, $G_i$ denote the coordinate frame fixed on the OCT scan head, camera, robot's gripper / end-effector (EEF). The $CW$ and $RW$ denote the ChArUco calibration pattern's and robot's world coordinate frame fixed on the calibration pattern and robot work station, respectively. $\mathbf{H}_{cg}$ denotes the transformation from $C_i$ to $G_i$, $\mathbf{H}_{og}$ denotes the transformation from $O_i$ to $G_i$, $\mathbf{H}_{gi}$ denotes the coordinate transformation from $G_i$ to the $RW$. We ignore the $i^{th}$ index for simplicity since the camera and OCT scan head is rigidly mounted on the gripper and they remains same through the experiment. For any given $i$ and $k$, $\mathbf{H}_{gik}$ denotes the coordinate transformation from $G_i$ to the $G_k$.

 \begin{figure*}
    \centering
    \subfloat[6D Scanning] {
        {\includegraphics[trim=0cm 0cm 0cm 0cm, clip, height=5cm]{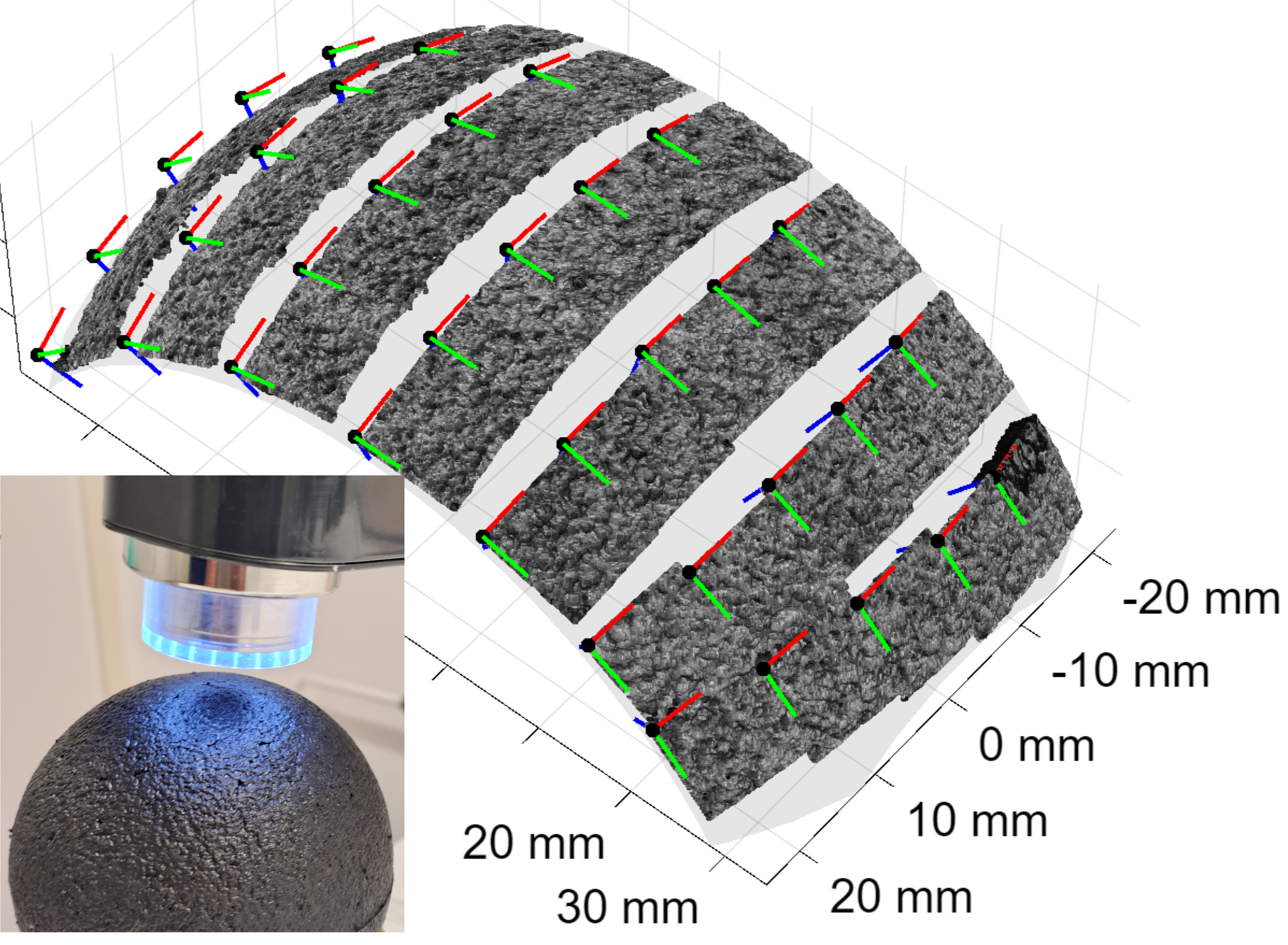}}          
        \label{img:6d_ball_surface_scanning}
   }
    \subfloat[3D-Translation vs. 6D Scanning] {
        \begin{tikzpicture}
            \node[inner sep=0pt] at (0cm, 0cm) {
            \includegraphics[trim=0cm 0cm 0cm 0cm, clip, height=5cm]{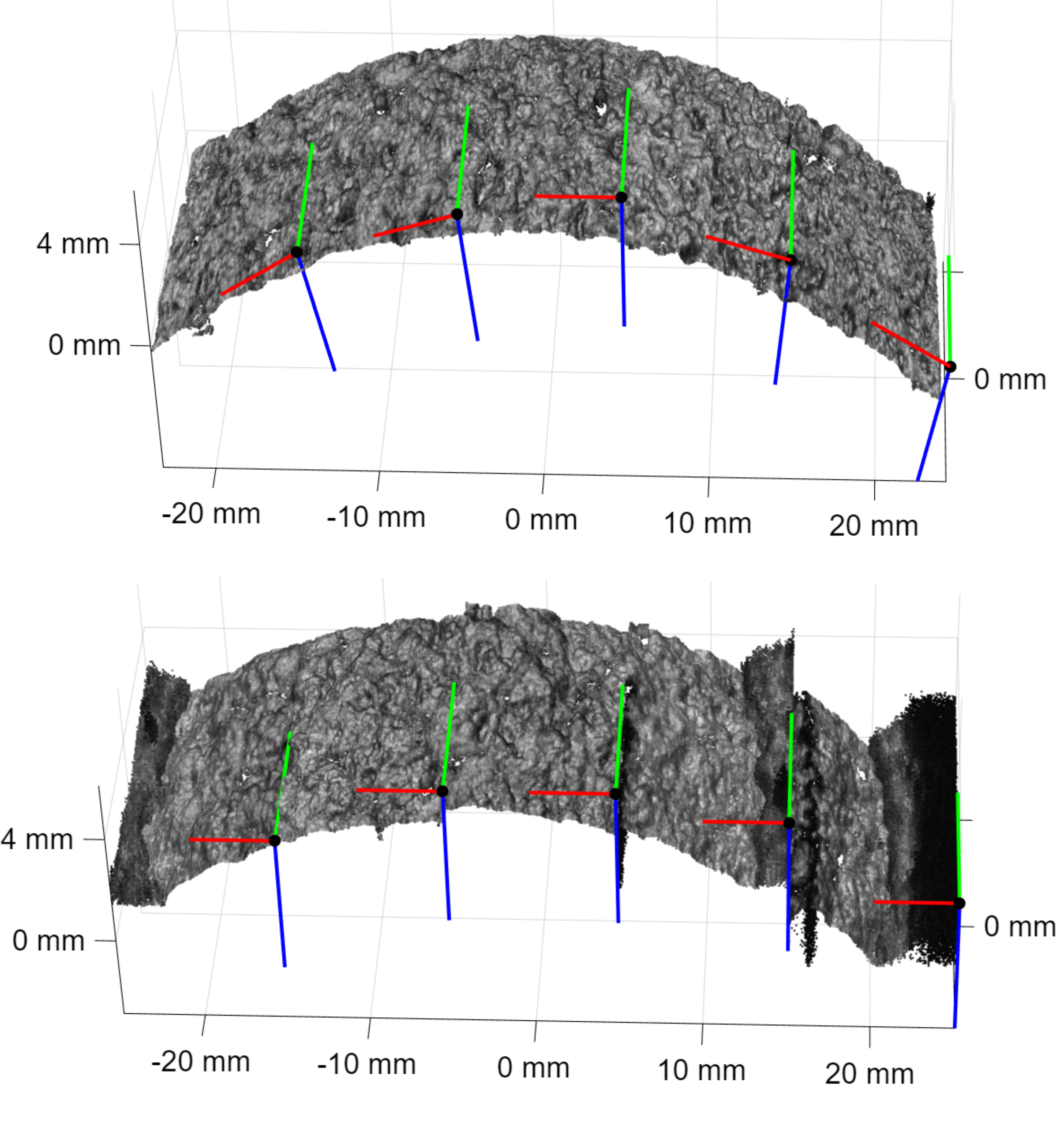}};
            \node[text width=30mm, align=center] at (0mm,26mm) {\scriptsize \textbf{6D Scanning}};
            \node[text width=30mm, align=center] at (0mm,-1mm) {\scriptsize \textbf{3D-Translation Scanning}};
            \draw[->, very thick, red] (3, -0.5) -- (1.75,-1) node[midway, above] {Artifact};
        \end{tikzpicture}   
        \label{img:3d_vs_6d_ball_surface_scanning}
   }
    \subfloat[Cross Section Surface] {
     {\includegraphics[trim=0cm 0cm 0cm 0cm, clip, height=5cm]{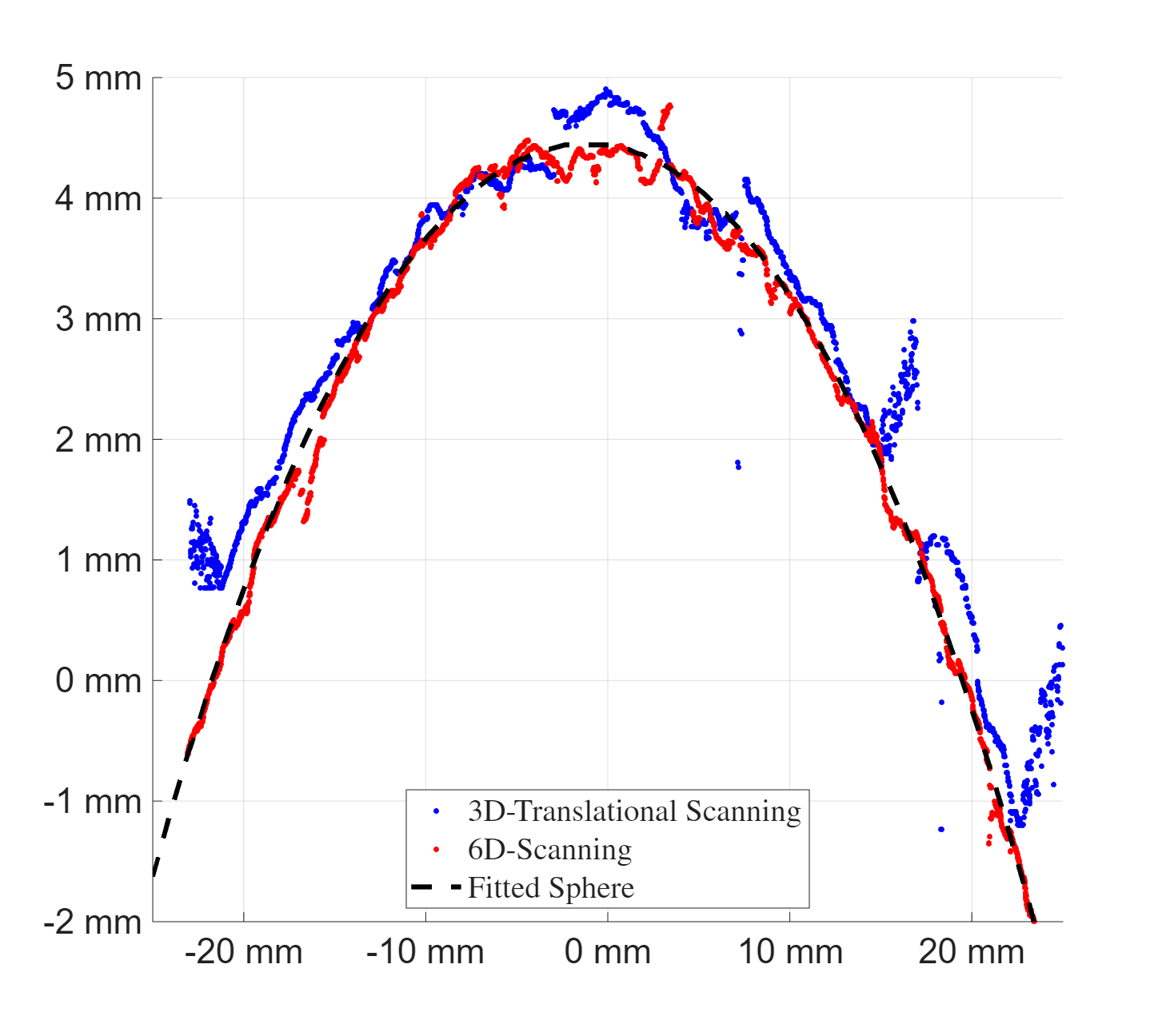}}     
        \label{img:LinePlot}  
     }
    \caption{\textbf{Spherical Phantom:} (a) 6D Surface stitching. (b) 6D surface scanning (top) and corresponding 3D-Translation with artifacts (bottom). (c) Extracted surface points from cross sections in plane 6D scanning (blue) and 3D-translation (red).}
 \end{figure*}
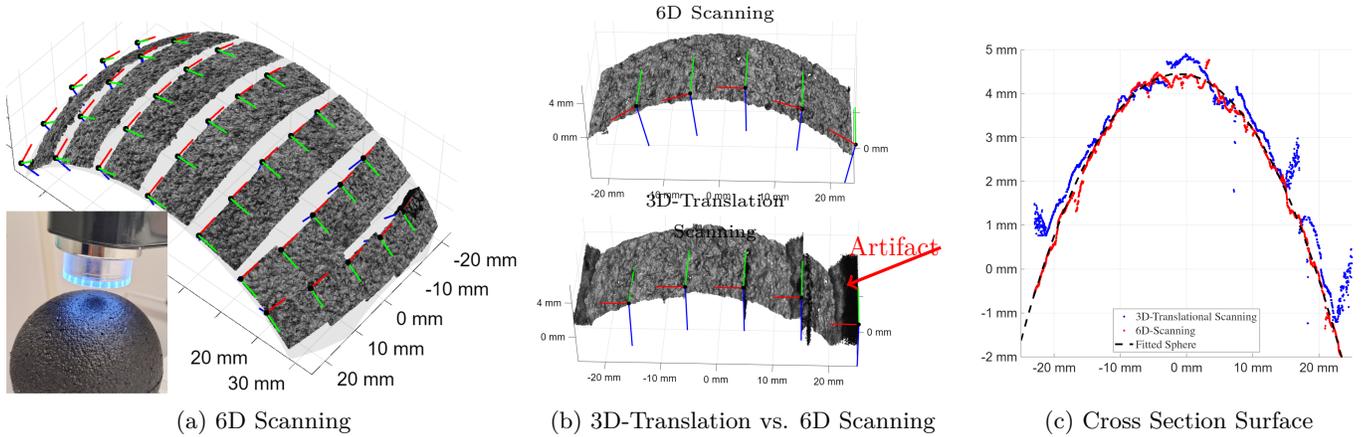 

\vspace{-3mm}
\subsection{Hand-in-Eye Calibration}
\label{sec:hand_in_eye_calibration}

Our task is to estimate the coordinate transformation from the camera $C_i$ and the OCT scan head $O_i$ to the robot's EEF $G_i$, i.e. $\mathbf{H}_{cg}$ and $\mathbf{H}_{og}$, using standard Hand-in-Eye calibration procedure, respectively. For this, we move the robot to $i \in N$ different configurations and compute the $\mathbf{H}_{gi}$ using the robot kinematics.

\noindent
\textbf{Camera-to-EEF} At each robot configuration $i$, when the calibration target is completely visible in the FOV of the camera, we acquire the rgb frame $\mathbf{I}^{rgb}_{i} \in \mathbb{R}^{H \times W \times 3}$. We detect the checkerboard squares and compute their corresponding world coordinates. Subsequently, the transformation $\mathbf{H}_{ci}$ between the calibration board world $CW$ and the camera frame $C_i$ is estimated using a Perspective-n-Point (PnP) solver based on 2D-3D point correspondence.

\noindent
\textbf{OCT-to-EEF} Fig.~\ref{img:aruco_marker_detection_workflow} summarizes our workflow to estimate $\mathbf{H}_{oi}$. We apply a median filter with a kernel size of \qtyproduct{3x3x3}{} to acquired OCT volumes $\mathbf{V}^{OCT}_{i} \in \mathbb{R}^{Z \times X \times Y}$. Secondly, we compute the 2D en-face projection along the depth axis to acquire the corresponding $\mathbf{I}^{OCT}_{i} \in \mathbb{R}^{X \times Y}$ image. We detect the ArUco marker ($j$) image points $[x^j,~y^j]^{\top}_{OCT}$ in $\mathbf{I}^{OCT}_{i}$ and reproject it back to original 3D volume to estimate the 3D ArUco corner positions $[x^j,~y^j,~z^j]^{\top}_{OCT}$. Finally, the transformation $\mathbf{H}_{oi}$ between the calibration board world $CW$ and the OCT scan head frame $O_i$ is estimated using a PnP solver based on 3D-3D point correspondence.

\vspace{-4mm}
\subsection{Calibration Accuracy Estimation}
\label{sec:calibration_accuracy}

Let, at $i^{th}$ and $k^{th}$ robot configuration, where $i \neq k$, $j^{th}$ ArUco marker is visible in both the camera image $\mathbf{I}^{rgb}_{i}$ and the OCT volume $\mathbf{V}^{OCT}_{k}$. We estimate the corresponding position of detected ArUco corner points $[x^j,~y^j,~z^j]^{\top}_{rgb}$ and $[x^j,~y^j,~z^j]^{\top}_{OCT}$ in camera image and OCT volume, respectively, relative to the robot world coordinate frame $CW$ and compute the reprojection error as follows,

\begin{align*}
    X_{CWi} = \mathbf{H}_{gi} ~ \mathbf{H}_{cg}  ~ \mathbf{H}_{ci} ~ [x^j,~y^j,~z^j~1]^{\top}_{rgb} \\
    X_{CWk} = \mathbf{H}_{gk} ~ \mathbf{H}_{og} ~ \mathbf{H}_{ok} ~ [x^j,~y^j,~z^j,~1]^{\top}_{OCT} \\
    \text{RMSE}_{CW} = || X_{CWi} - X_{CWk}||_2
    \tag{1}\label{eq:reprojection_error}
\end{align*}

where, the poses $i \in 1 \cdots N$ and $k \in 1 \cdots M$ at which the camera image and OCT volume are acquired can be different, and $N$, $M$ are number of robot configurations with camera-robot and oct-robot pairs, respectively.

\vspace{-3mm}
\subsection{Surface Scanning}

We compare our \textbf{proposed 6D scanning} approach to a \textbf{conventional 3D-translation scan} by moving the robot with and without orientation change during consecutive scans. Based on the detected surface depth we adjust the axial-position of the robot with fixed lateral xy-increments. We evaluate our scanning approach on a styrofoam sphere with a radius of approximately \SI{50}{\mm}. We acquire $35$ OCT volumes while rotating the OCT scan head around the center of the sphere, as depicted in Fig.~\ref{img:6d_ball_surface_scanning}. Poses are refined manually to capture sufficient surface information. During 3D-translation $5$ consecutive scans of the sphere are acquired, refer Fig.~\ref{img:3d_vs_6d_ball_surface_scanning}. Additionally, we perform surface scans on a chicken phantom by acquiring $65$ and $32$ volumes during 6D scanning and 3D-translation scanning, respectively. The scan positions are indicated in Fig.~\ref{fig:ChickenPhantom}. We detect the surface in scanned OCT volumes by estimating the maximum intensity along each scan line. We convert surfaces to colored point clouds for visualization.




\vspace{-3mm}
\section{Results}

\begin{figure}
  \centering
  \begin{tikzpicture}
    \node[inner sep=0pt] at (0cm, 0cm) {
    \includegraphics[trim=0cm 0cm 0cm 0cm, clip, width=\linewidth]{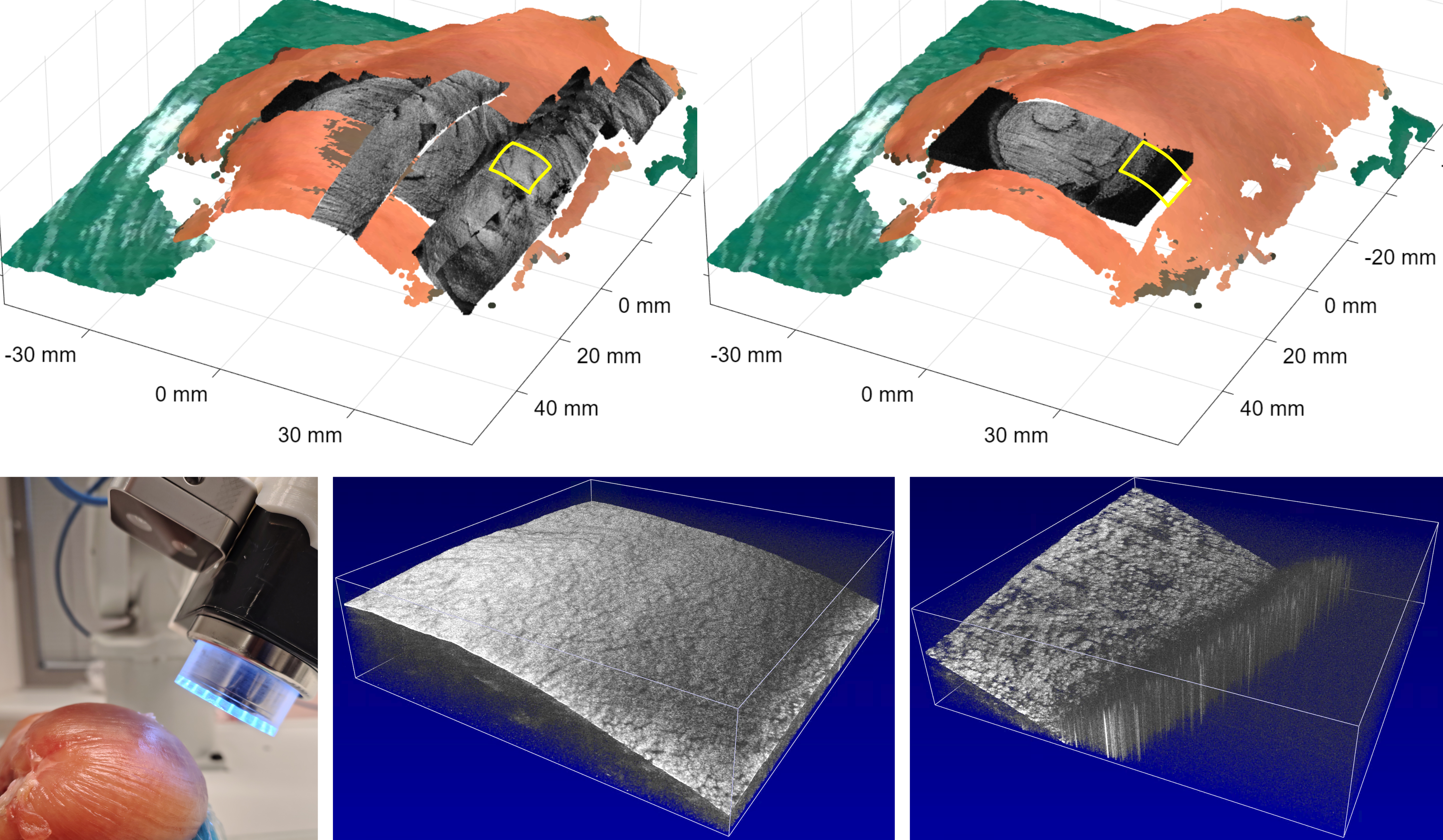}};
    \node[text width=30mm] at (-15mm,29mm) {\scriptsize \textbf{6D Scanning}};
    \node[text width=30mm] at (25mm,29mm) {\scriptsize \textbf{3D-Translation Scanning}};

    \draw [fill=black] (-12mm,15mm) circle (0.2mm); 
    \draw [] (-12mm+5mm,15mm-3mm) -- (-12mm,15mm);
    \node[text width=5mm] at (-12mm+8mm,15mm-3mm) {A};

    \draw [fill=black] (-20mm+45mm,15mm) circle (0.2mm); 
    \draw [] (-20mm+45mm-5mm,15mm-5mm) -- (-20mm+45mm,15mm);
    \node[text width=5mm] at (-20mm+45mm-5mm,15mm-5mm) {B};

    \node[text width=5mm, color=white] at (-20mm,-23mm) {A};
    \node[text width=5mm, color=white] at (-20mm+35mm,-23mm) {B};
 \end{tikzpicture}
  \caption{\textbf{Soft-Tissue Phantom:} 6D and 3D-translation scans acquired on chicken tissue of OCT volumes and RGB-D data. The positions of two exemplary OCT volumes are indicated as A and B.}
 \label{fig:ChickenPhantom}
\end{figure}
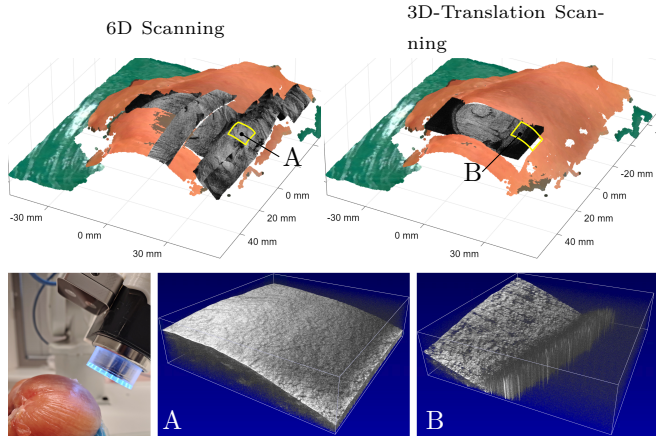

\noindent
First, we evaluate the consistency of our hand-eye calibration. Table~\ref{tab:OCT_hand_in_eye_calibration} shows the mean and standard deviation for $\mathbf{H}_{cg}$ and $\mathbf{H}_{og}$ estimates for 3 calibration runs. We report a maximum standard deviation of \SI{0.4}{\milli\meter} and 0.24~deg for translation and orientation, respectively. Second, we study the calibration accuracy. We report the reprojection error for different numbers of OCT volumes utilized during calibration, depicted in Fig.~\ref{img:translationError_Calibration}. It stands out that the error reaches a plateau after 25 volumes. We report a mean reprojection error of \SI{0.36(0.25)}{\milli\meter}. Depicted in Fig.~\ref{img:distanceError_Calibration} is the reprojection error estimated on ArUco markers, which were not included in the calibration data set. The error increases slightly for ArUco markers at a larger distance from the center of the ChArUco board, and we report a maximum error of \SI{0.53(0.13)}{\milli\meter}. Third, we perform surface scans on phantoms. Fig.~\ref{img:6d_ball_surface_scanning} depicts the 6D surface scan of the spherical phantom. We fit a sphere to the surface scans and report a mean deviation error between the OCT surface points and the fitted sphere of \SI{0.13(0.12)}{\milli\meter}. \replaced{Further, we compare conventional 3D‑translation scanning to our 6D scanning approach (Fig.~\ref{img:3d_vs_6d_ball_surface_scanning}): the error in estimated sphere radius is \SI{8.26}{\milli\meter} for 3D translation versus \SI{0.35}{\milli\meter} for 6D scanning.} {
Further, we compare the conventional 3D-translation scanning to our 6D scanning approach, as depicted in Fig.~\ref{img:3d_vs_6d_ball_surface_scanning}.}\deleted{add a quantitative metric.} Please note, OCT volumes containing a steeper surface relative to the OCT imaging reference frame acquired during 3D-translation scanning have artifacts. The partially incomplete surface scans acquired during 3D-translation scanning can also be seen in Fig.~\ref{img:LinePlot}. Moreover, we compare qualitatively both scanning approaches on a chicken phantom. Depicted in Fig.~\ref{fig:ChickenPhantom} is merged the RGB-D and OCT volumetric data. It stands out that 3D-translational scanning fails for larger surface curvatures while 6D scanning provides reliable surface information.



\begin{table}[!t]
\footnotesize
\caption{\textbf{Hand-in-Eye Calibration Results} Camera (top row) and OCT scan head (bottom row)}
  \begin{tabular*}{\linewidth}{lccccccc}
    \hline  
    &  &  \multicolumn{3}{c}{Position $(\SI{}{\mm})$} &  \multicolumn{3}{c}{Orientation $(\SI{}{\deg})$}   \\  \cmidrule{3-5} \cmidrule{6-8}
    &  & x & y & z & r & p & y \\ \midrule
    \multirow{ 2}{*}{$\mathbf{H}_{cg}$} &$ \mu$ & $72.33$ & $-62.51$ & $85.3$ & $68.76$ & $70.41$ & $67.71$ \\
            & $\sigma$  & $0.38$ & $0.40$ & $0.12$ & $0.13$ & $0.06$ & $0.04$ \\
    \hline
    \multirow{ 2}{*}{$\mathbf{H}_{og}$} & $\mu$ & $140.28$ & $-9.40$ & $92.15$ & $0.27$ & $91.55$ & $-0.42$ \\
            & $\sigma$  & $0.22$ & $0.04$ & $0.19$ & $0.12$ & $0.24$ & $0.19$ \\
    \hline
  \end{tabular*}
\label{tab:OCT_hand_in_eye_calibration}
\end{table}

\vspace{-3mm}
\section{Discussion and Conclusion}
We present a 6D calibration pipeline for estimating the transformation between the OCT scan head and a RGB-D camera mounted to a robot gripper. Our calibration results are consistent over multiple experiment iterations and enable consistent 6D scanning of large curved surfaces with fewer stitching artifacts, which typically limits conventional 3D translation scanning approaches, as indicated by our results. The full calibration of RGB-D images to OCT volumetric data allows wide range of robot orientation configurations during scanning of curved tissue surfaces and enables collision-free optimal scanning trajectories with a robot. \added{Limitations: evaluation was restricted to phantom data; future work will target clinical validation (in‑ and ex‑vivo) and benchmarking against conventional calibration methods.} \deleted{Limitation: more comparison to other methds,  validation on in-vivo or ex-vivo clinical data would have been preferrable, "comparison to some other calibration or registration-based scanning strategy", which?, marker placement or robustness to partial marker visibility.}

\vspace{-3mm}
\section{Acknowledgments}

This research was co-funded by the MARLOC project (DFG, grant SCHL 1844-10-1) and by the European Union under Horizon Europe programme grant agreement No. 101059903; and by the European Union funds for the period 2021-2027. Conflict of interest: none. Informed consent: obtained. Ethical approval: not applicable.


\vspace{-3mm}
\bibliographystyle{IEEEbib}
\bibliography{strings,refs}

\end{document}